# EMBEDDED COMPUTER VISION SYSTEM APPLIED TO A FOUR-LEGGED LINE FOLLOWER ROBOT

**Beatriz Arruda Asfora**
Universidade Federal de Pernambuco, Av. Professor Morais Rego, 1235 - Cidade Universitária, Recife - PE, 50670-901.
b.asfora@hotmail.com

*Abstract. Robotics can be defined as the connection of perception to action. Taking this further, this project aims to drive a robot using an automated computer vision embedded system, connecting the robot's vision to its behavior. In order to implement a color recognition system on the robot, open source tools are chosen, such as Processing language, Android system, Arduino platform and Pixy camera. The constraints are clear: simplicity, replicability and financial viability. In order to integrate Robotics, Computer Vision and Image Processing, the robot is applied on a typical mobile robot's issue: line following. The problem of distinguishing the path from the background is analyzed through different approaches: the popular Otsu's Method, thresholding based on color combinations through experimentation and color tracking via hue and saturation. Decision making of where to move next is based on the line center of the path and is fully automated. Using a four-legged robot as platform and a camera as its only sensor, the robot is capable of successfully follow a line. From capturing the image to moving the robot, it's evident how integrative Robotics can be. The issue of this paper alone involves knowledge of Mechanical Engineering, Electronics, Control Systems and Programming. Everything related to this work was documented and made available on an open source online page, so it can be useful in learning and experimenting with robotics.*

## 1. INTRODUCTION

"Robotics is the intelligent connection of perception to action" (Brady, 1984). A robot is then characterized as a mechanical and articulated device that can obtain information from the environment using sensors, make decisions based on this information and previous settings, and interact with the surroundings using actuators. Cameras are sensors commonly used on robots, and the field of Computer Vision addresses the problem of emulating the human vision and interpreting the 3D world based on 2D images. Computer Vision deals not only with the image itself, but foremost with the logical act of assuming facts about the world and using known models to infer what that image represents. (Szeliski, 2010).

This paper presents different Computer Vision techniques applied to line following, using camera as a sensor to get data and ultimately control the robot's path, and therefore intelligently connect perception to action. It also shows that is possible to use a simple and low cost robot to learn and experiment with Robotics, which involves knowledge in Mechanics, Electronics, Control Systems and Programming.

Florêncio and Baltar (2012) first developed the structure of the four-legged robot here described, but for other purposes and without an embedded computer vision system. That means the robot was dependent of the image processing capabilities of a desktop computer. This project differs from (Florêncio e Baltar, 2012) by taking a step further and bringing the image processing and decision-making to the robot itself, keeping the constraints of simplicity, replicability and financial viability.

Line following revolves around one basic problem: segmentation of the line from the background so the robot can follow the delimited path. Although the problem may be simple, it addresses some controlling and computer vision challenges present in situations that are more complex. For instance, the system needs to be relatively robust when dealing with changes in brightness, and fast enough so the robot is able to follow a set point that changes continuously as the device moves.

Three approaches were considered when implementing the four-legged robot as a line follower. They use the same robotic structure, but different image capturing tools and image processing and analysis methods. The first and second ones use Android Platform to capture the image and Processing to implement the control logic, comparing Otsu's Bimodal Limiarization Method (Otsu, 1979) with thresholding based on color combinations found through experimentation. The third one uses Pixy camera and a color tracking method in HSV color space, whereas the control logic is implemented directly at the Arduino Platform.

This text is organized as follows. Section 2 presents the project methodology: the mechanical project of the robot, as well as the prototype, actuators, image and programming tools chosen. Section 3 addresses the first and second approaches (Android, Processing & Thresholding), Section 4 addresses the third one (Pixy, Arduino & Color Tracking in HSV), analyzing its results, limitations and advantages. Section 5 presents the conclusion.

## 2. PROJECT AND MATERIALS

### 2.1 Open Source Tools

#### 2.1.1 Android Platform

Android is a mobile operating system launched in 2003 and is currently present in over one billion mobile cellphones around the world. It is based on the Linux system and Java programming language, and it is highly customized. In order to develop apps for Android, one can download the open source Software Development Kit (SDK) at (Developers), which contains the necessary material for the different Android versions, or APIs (Application Programming Interface). In this project the API 8 (Android 2.2.2), API 15 (Android 4.0.4) and API 16 (Android 4.1.2) were tested.

#### 2.1.2 Processing

Ben Fry and Casey Reas developed Processing in 2001 with the goal of being an open source software development tool for both professionals and new programmers. Processing offers a collaborative online community of users, with libraries, codes, tutorials and free download of its development environment at (Processing 2.0). Also based on Java language, Processing is recommended for Android apps development since it has an Android mode itself. For this project, Processing versions 1.5b, 2.0, 2.1, and 2.1.2 were tested.

Processing founders indicate the Ketai Library for Processing Android to access and work with Android sensors, including image digitalization for the image captured by the Android camera. Ketai versions v9 and v8 were tested.

#### 2.1.3 Arduino Development Platform

Arduino is an open-source electronics platform based on easy-to-use hardware and software. It has a microcontroller chip that functions as the board's brain, and several complementary components that make it easier to connect with an external circuit. Both the software and hardware of Arduino are open source and its IDE (Integrated Development Environment) can be downloaded at the website in (Arduino). This project uses the Arduino Duemilanove Bluetooth, which is based on the ATmega328 microcontroller and has 14 digital input/output pins (of which 6 can be used as PWM outputs), 6 analog inputs, a 16 MHz crystal oscillator, screw terminals for power, an ICSP header, and a reset button.

The Bluetooth module on this Arduino is useful when receiving data from the Android. The Android device can capture and process the image and calculate what the robot should do, and then send the command to the Arduino via Bluetooth.

#### 2.1.4 Pixy Camera

Pixy is the latest product of a long line of Charmed Labs cameras, called CMUcams. Charmed Labs invests in easy-to-use and low cost technology. According to one of the developers: "Our vision system is designed to provide high-level information extracted from a camera image to an external processor that may, for example, control a mobile robot" (Rowe, 2002).

The main problem with image sensors is they produce a large amount of data and overload the processor, making it difficult to deal with anything else. Pixy addresses this problem by pairing a microprocessor to the camera, dedicated to obtain and organize the data in order to send only the central information about the tracked object: its signature (color), x and y centers, width and height. Pixy's firmware makes it simple for the user to choose the color that should be tracked, just by pressing a button or indicating the desired color on Pixy's development environment (CMUcam, 2013).

Pixy's frame rate is 50Hz and at each frame it is capable of tracking up to 135 objects. The device sends the information via UART serial, SPI, I2C, USB or Digital I/O to the platform it is connected. For this project Pixy connects to Arduino through SPI port. Although Pixy digitalizes the image in RGB format, it converts it to HSV for image analysis that will be detailed in Section 4 of this paper.

### 2.2 The Four-Legged Robot

The whole cost of the robot prototype, without tax or delivery fees, is approximately U$80.00, without Pixy Camera, or U$149.00, with Pixy Camera. The robot is therefore financially viable when compared to other robotic platforms. This section also shows the robot's simplicity and, since everything related to this work was documented and made available on an open source online page[1], its replicability.

The robot's mechanical structure is shown in Fig 1. It is composed by eight servomotors model HK15138 (in orange), 3mm thick aluminum plates at various sizes (in gray) and screws/nuts for fixation.

---
[1] http://robolivre.org/conteudo/embedded-computer-vision-system-applied-to-a-four-legged-line-follower-robot

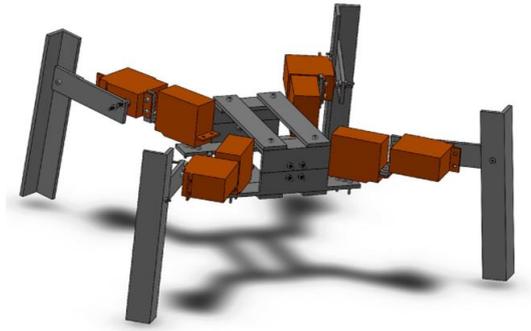

Figure 1. Mechanical structure of the four-legged robot

The Arduino Duamilanove Bluetooth, through its PWM pins, controls the servomotors. The servo's goal is to transform an electric sign in proportional movement. For this, it uses a potentiometer as a sensor of the motor axis' position, comparing the potentiometer resistance to the sign received by Arduino and activating the motor to correct the difference between them if needed. In other words, the duration of the pulse applied to the servomotor controls its rotation angle. The robot is powered by a Lipo NanoTech 1.0 12V battery and a power distribution board, which also steps down the voltage to 6V to meet servo parameters.

The robot translates and rotates by activating the servos sequentially, in a way that there are always two legs on the floor to maintain the balance. Consider the four servos connected to the center as the shoulders and the other four as the elbows, where each pair of elbow and shoulder belongs to one quadrant of the coordinate plane. The camera points to the positive *x axis*, as shown in Fig. 2.

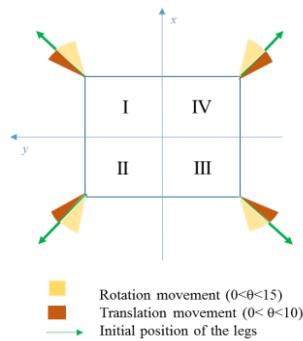

Figure 2. Robot's movement schematic

To move one of the robot's legs first the shoulder contracts towards the *y axis* for 200ms. Then it starts to extend towards the *x axis* while the elbow contracts lifting the leg for 100ms. The elbow then extends, placing the leg on the floor while the shoulder finishes extending for another 100ms. Every move is a combination of legs steps. To move forward for example, the shoulders in quadrants I and II (as well as III and IV) are synchronized, while the elbows of quadrants I and III (as well as II and IV) are synchronized, lifting diagonally opposed legs at the same time to keep the balance.

All the algorithms developed in this project use the line center to decide what the robot should do next. If the line center is aligned with the robot's vision, then it moves forward. If it is to the right, then the robot turns right before moving. The optimum angle range of the servos for each step were found to be 10 degrees for translation and 15 degrees for rotation, because for higher ranges the robot tends to deviate too much from the path.

## 3. ANDROID, PROCESSING AND THRESHOLDING

In this approach, the line is 25mm thick and of color black, while the background is white. The limiarization goal is to separate the image in foreground (region of interest) and background by choosing a threshold. The transformation function follows:

$$g(x,y) = \begin{cases} l_1 \text{ if } f(x,y) \leq T \\ l_2 \text{ if } f(x,y) > T \end{cases} \qquad (1)$$

Where $f(x,y)$ is each point of the image function, $I_1$ and $I_2$ are the new intensity values for each region (0 and 1 for binary images), $T$ the threshold and $g(x,y)$ each point after the transformation. This segmentation technique assumes that: pixel intensities are different in different regions and are similar within the same region.

To establish the basis for comparison, the popular Otsu Method is applied to identify the line. Otsu's Bimodal Limiarization is a "nonparametric and unsupervised method of automatic threshold selection (…). An optimal threshold is selected by the discriminant criterion, namely, so as to maximize the separability of the resultant classes in gray levels" (Otsu, 1979). Let the pixels of a given image being represented in $L$ gray levels $[1,2…,L]$, the histogram of this image gives the total number of pixels at each level. The pixels can be separated in two classes $C_1$ and $C_2$, where $C_1$ contains the pixels with intensity up until level $k$ and $C_2$ contains the pixels from level $k+1$ until level $L$. The probability of a pixel being in one of these classes can be calculated, as well as the weighted mean and variance of each region, and weighted mean and variance for the global image. By iteratively testing all levels, Otsu's optimum threshold finds the $k$ level that maximizes the variance between classes $C_1$ and $C_2$.

To apply this method however, first the image needs to be converted to gray levels since Android digitalizes the image in RGB format. Consider the image captured by the robot's vision system at a given time in Fig. 3a, its conversion to gray levels in Fig. 3b and the binary image after applying Otsu's optimum threshold in Fig. 3c:

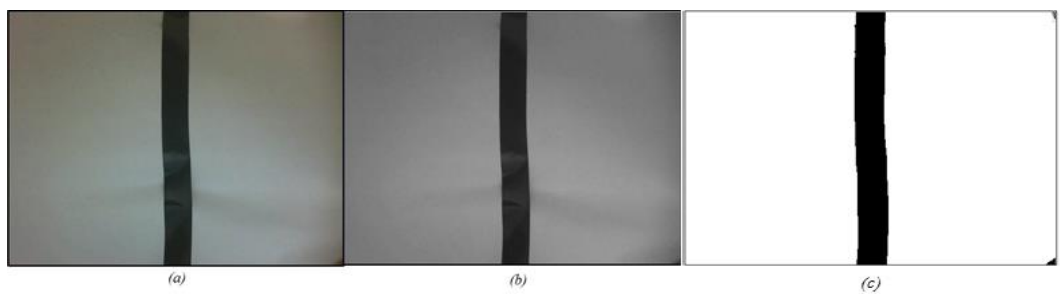

Figure 3. (a) original image; (b) image in gray levels; (c) binary image after applying Otsu Method

Otsu Method successfully identifies the line and presents few misidentified foreground regions near the image border, but it needs to be repeated at every frame: convert to gray scale, compute the histogram and iteratively find the optimal threshold.

The other method tested for this application is an experimental thresholding based on RGB color combinations, which also aims to maximize the separability of different regions. The RGB values for 200 samples of visually black and white points were collected in three scenarios of gradually increased brightness, to simulate changes on the environment while the robot is following the path. By testing and comparing several combinations, it was possible to choose one that would correctly distinguish the black and white regions, classifying each pixel accordingly even in scenarios with less light. Table 1 shows the results of this technique using the combination shown below:

*Black if $R+G+B < 250$; $G-B < 30$; $R-B > -30$* (1)

Table 1. Results for experimental thresholding in RGB

|  | Black point | | | White | | | |
| --- | --- | --- | --- | --- | --- | --- | --- |
| Scenarios[1] | 1 | 2 | 3 | 1 | 2 | 3 | TOTAL |
| Collected samples | 50 | 50 | 50 | 50 | 50 | 50 | 250 |
| Properly classified | 50 | 45 | 47 | 50 | 50 | 50 | 242 |
| Improperly classified | 0 | 5 | 3 | 0 | 0 | 0 | 8 |
| Success rate | 100% | 90% | 94% | 100% | 100% | 100% | 97% |

[1] Scenario 1 is the darkest and scenario 3 is the brightest

Figure 4 shows the binary image resultant from this technique when applied to Fig.3a:

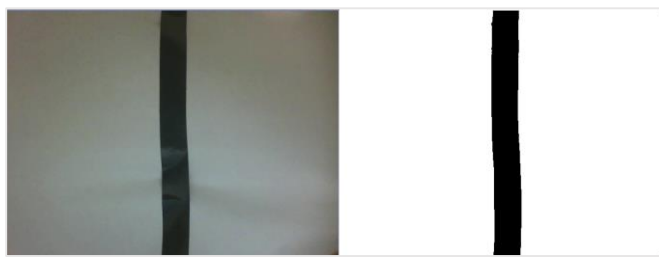

Figure 4: Original image (left) and binary image using experimental thresholding (right)

In a visual comparative analysis, the second technique shows a similar performance to Otsu Method for this application. The experimental thresholding described here has the advantage of being more intuitive and requiring less computational effort, since it needs only three comparison operations until it can identify in which region the pixel belongs.

In order to identify the line center, the program loops for one line of the image and checks for black pixels. Then it calculates the mean of the black pixels located, in other words, the sum of the *x* coordinates of the black pixels divided by the number of black pixels found. The decision of where to move next is based on where the line center is in relation to the robot's vision center, as described in Section 2 of this paper. The fixed *y* coordinate for the loop is experimentally chosen as 1/6 of the image height, to avoid being too close to the border but still in the beginning of the robot's view so it has time to react. Figure 5 demonstrates the location of the line center (in red) and the binary image using the experimental thresholding method. For practical implementation, it was found that Processing 2.1.2, Android 4.1.2 (API 16) and Ketai v9 are compatible, while the older versions are not.

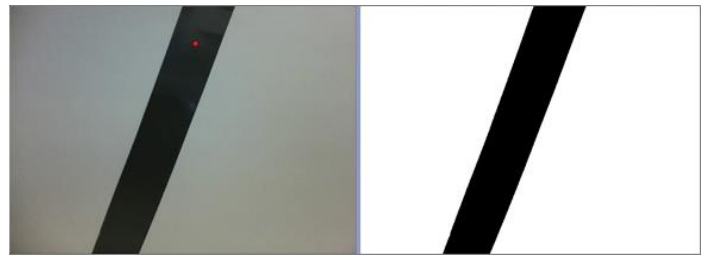

Figure 5: Location of the line center (in red) and binary image using the experimental thresholding method (right)

## 4. PIXY, ARDUINO AND COLOR TRACKING IN HSV

Pixy, or CMUcam5, uses techniques that are simple, creative and require less computational effort than traditional ones such as Otsu, so it is possible to implement them in a memory limited and low cost hardware - a microprocessor. The camera uses a CMOS image sensor, which means that every electric signal generated by each photodiode is amplified, converted and digitalized at the same point it is registered. This gives Pixel an advantage because data can be processed and computed while still forming the image: in between pixels, at the end of a line of pixels or at the end of a frame, with little or no storage so not to overwhelm the microprocessor.

Pixy uses multi-thresholding in two, hue and saturation, of the three channels of the HSV space to detect regions of interest that differ in color from the surroundings. It does not use the brightness (V) so the system is more robust regarding changes in lighting and exposure.

A region is defined by four thresholds: an upper hue threshold, lower hue threshold, upper saturation threshold and lower saturation threshold. This means four integer comparison operations are needed for each pixels:

$$\text{if } H \geq lowerhue \text{ AND } H \leq upperhue \text{ AND } S \geq lowersat \text{ AND } S \leq uppersat. \qquad (2)$$

To do these comparisons more efficiently and using less memory, *Pixy* applies an adaptation of the method described in (Bruce, 2000). The color space is decomposed in a boolean way and stored in matrixes, making it possible to compare bits instead of integers. Each region is the product of two functions, each in a color channel, and the comparison operation is made element by element of the matrix. For example, decomposing the channels in 10 levels, a given color has the following matrixes *HClass[] = {0,1,1,1,1,1,1,1,1,1} and SClass[] = {0,0,0,0,0,0,0,1,1,1}*. Consider a pixel has the value of *HS(1,8)*, then the following operation confirms the pixel is on both classes and therefore is of interest:

$$HClass[1] \text{ AND } SClass[8] = 1 \text{ (true)} \qquad (3)$$

To establish the thresholds for comparison, the user selects the region of interest at Pixy's development environment and based on the image coordinates Pixy identifies the wrapped pixels and stores the levels values for each channel (thresholds). Pixy scans the image pixel by pixel from the top left corner to the bottom right corner, deciding if each pixel is of interest or not. The upper and lower coordinates of each region of pixels of interest is stored forming rectangular "delimiting boxes". At each new pixel classified as foreground, the box expands to include the new pixel. The software maintains an iterative count of the number of pixels found and also their *(x,y)* coordinates, and at the end of each frame it calculates the box centroids. The height and width of the regions are also available since the boxes coordinates are stored. Pixy also applies some simplifications of statistical functions and filters, such as only considering the pixel as part of the foreground region only if the number of consecutive pixels before it is higher than a set value, avoiding image noise. As currently there is no official publication of Pixy's methodology, CMUcam4 documentation and Pixy's open source firmware are the basis for the techniques described here.

In order to implement Pixy on the four-legged line follower robot, the color of the path was chosen as red. Since Pixy uses thresholding in HSV space, black was not a suitable color because it only differs from white due to brightness – which is not a parameter used in Pixy's method. The recommended colors for using Pixy are red, orange, yellow, green, cyan, blue and violet. For each different color chosen by the user, Pixy calls it a signature from one to seven. During implementation, two different regions of the path were set as signatures prevent errors.

Pixy sends the relevant data of each region to Arduino in matrix form, where the first matrix element is the data (signature, x center, y center, height and width) of the biggest region found. Since the goal is to follow a line, the biggest region is the closest one to the robot, so the robot's next action must respond to this data. Similarly to what was described in Section 2, the robot's action to move forward, left or right is based on the location of the line $x$ center in relation to the robot's vision center.

The following limits showed good results when tested (Pixy's *x axis* ranges from 0 to 399):

$$Move \begin{cases} left\ if\ x' \leq 130 \\ foward\ if\ 130 < x' \leq 170 \\ righ\ if\ x' > 170 \end{cases} \quad (4)$$

Where *x'* is the closest region *x* center.

As said before, Pixy is capable of offering new data every 20ms, which is 5 times faster than the robot is capable of moving, because each step takes 200ms. Therefore the routine implemented for line following gets data from Pixy, makes a decision about what the robot should do and waits for the robot to take one step in the chosen direction before getting new data. If no region of interest is found, the robot stops and gets new data until Pixy is able to recognize the tracked color. Figure 6 shows the robot during one of the tests:

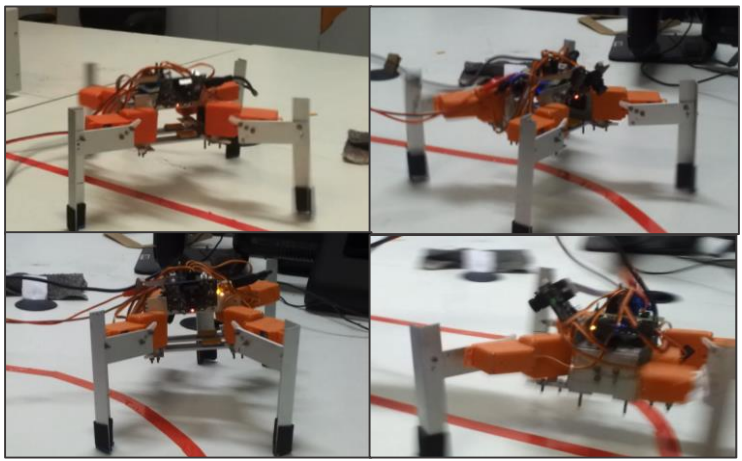

Figure 6: The four-legged line follower robot paired using Pixy camera to follow a delimited path

Paired with Pixy camera, the robot is capable of correctly following the path if it is a straight line, smooth turn and also if it is a sharp turn, with an average velocity of 0.03m/s.

## 5. CONCLUSION

This paper portrays the use of Computer Vision to control the trajectory of an autonomous four-legged robot applied on a typical mobile robot issue: line following. It analyzes the vision issue through different approaches. By using the traditional Otsu Bimodal Limiarization Method as the comparative basis for the experimental thresholding based on color combinations developed, it concludes the later presents results as good as the Otsu Method and is more suitable for this application. Then a third approach is investigate and chosen for practical tests with the robot platform. It uses the creative technique of an innovative tool, Pixy Camera, that track colors based on levels of hue and saturation, representing a more robust system to changes in light and exposure. Pixy represents the best fit for the robot platform used, since it also works with the constraints of simplicity, financial viability and replicability.

It is important to point out that despite of the relative simplicity of the line following issue, it enables complex approaches and creative solutions, contributing on learning and experimenting with Robotics.

## 7. RESPONSIBILITY NOTICE

The author is the only responsible for the printed material included in this paper.